\documentclass[lettersize,journal]{IEEEtran}
\usepackage{amsmath,amsfonts}
\usepackage{algorithmic}
\usepackage{algorithm}
\usepackage{array}
\usepackage{textcomp}
\usepackage{stfloats}
\usepackage{subcaption}
\usepackage{url}
\usepackage{verbatim}
\usepackage{graphicx}
\usepackage{cite}
\begin{document}

\title{LLM-MARS: Large Language Model for Behavior Tree Generation and NLP-enhanced Dialogue in Multi-Agent Robot Systems}


\author{
Artem Lykov,
Maria Dronova,
Nikolay Naglov,
\\
Mikhail Litvinov,
Sergei Satsevich,
Artem Bazhenov,
\\
Vladimir Berman,
Aleksei Shcherbak and
Dzmitry Tsetserukou,~\IEEEmembership{Member,~IEEE}
\thanks{\textbf{©2023 IEEE. This work has been submitted to IEEE for possible publication. Copyright may be transferred without notice, after which this version may no longer be accessible}}
\thanks{The authors are with the Intelligent Space Robotics Laboratory, Center for Digital Engineering, Skolkovo Institute of Science and Technology (Skoltech), 121205 Moscow, Russia. {\tt\small \{Artem.Lykov, Maria.Dronova, Nikolay.Naglov, Mikhail.Litvinov2, Sergei.Satsevich, Artem.Bazhenov, Vladimir.Berman, Aleksei.Shcherbak, D.Tsetserukou\}@skoltech.ru\ }}%
}


\markboth{Lykov \MakeLowercase{\textit{et al.}}}%
{Lykov \MakeLowercase{\textit{et al.}}: LLM-MARS: Large Language Model for Behavior Tree Generation and NLP-enhanced Dialogue in Multi-Agent Robot Systems}




\maketitle

\begin{abstract}
  This paper introduces LLM-MARS, first technology that utilizes a Large Language Model based Artificial Intelligence for Multi-Agent Robot Systems. LLM-MARS enables dynamic dialogues between humans and robots, allowing the latter to generate behavior based on operator commands and provide informative answers to questions about their actions. LLM-MARS is built on a transformer-based Large Language Model, fine-tuned from the Falcon 7B model. We employ a multimodal approach using LoRa adapters for different tasks. The first LoRa adapter was developed by fine-tuning the base model on examples of Behavior Trees and their corresponding commands. The second LoRa adapter was developed by fine-tuning on question-answering examples. Practical trials on a multi-agent system of two robots within the Eurobot 2023 game rules demonstrate promising results. The robots achieve an average task execution accuracy of 79.28\% in compound commands. With commands containing up to two tasks accuracy exceeded 90\%. Evaluation confirms the system's answers on operators questions exhibit high accuracy, relevance, and informativeness. LLM-MARS and similar multi-agent robotic systems hold significant potential to revolutionize logistics, enabling autonomous exploration missions and advancing Industry 5.0.
\end{abstract}


\section{Introduction}
\IEEEPARstart{I}{n} recent years, remarkable advancements have been made in the realms of robotics and artificial intelligence. Various methodologies have emerged to enhance the logical capabilities of robotic systems, with a particularly intriguing avenue of research involving the integration of large language models (LLMs), like GPT. These LLMs possess the exceptional ability to break down complex natural language tasks into simpler sub-tasks that can be executed by robots. Additionally, LLMs have been leveraged to develop human-robot interaction systems using natural language processing (NLP) techniques. 

To address a range of challenges, multimodal solutions have also been explored. The emergence of open LLMs has marked a turning point in LLM research, as they can be openly trained on custom data, making them accessible and capable of performing instruction-following tasks. However, only limited research has focused on using LLMs to generate entire robot behaviour trees (BTs) that simultaneously consider the handling of a large number of possible scenarios. Such models are specifically designed to leverage natural language processing techniques to comprehend human speech and generate intricate operating sequence for robots based on the acquired information. This approach exhibits significant potential in enhancing the adaptability and flexibility of artificially intelligent robot systems. 

\begin{figure}
    \centering
    \includegraphics[width=\linewidth]{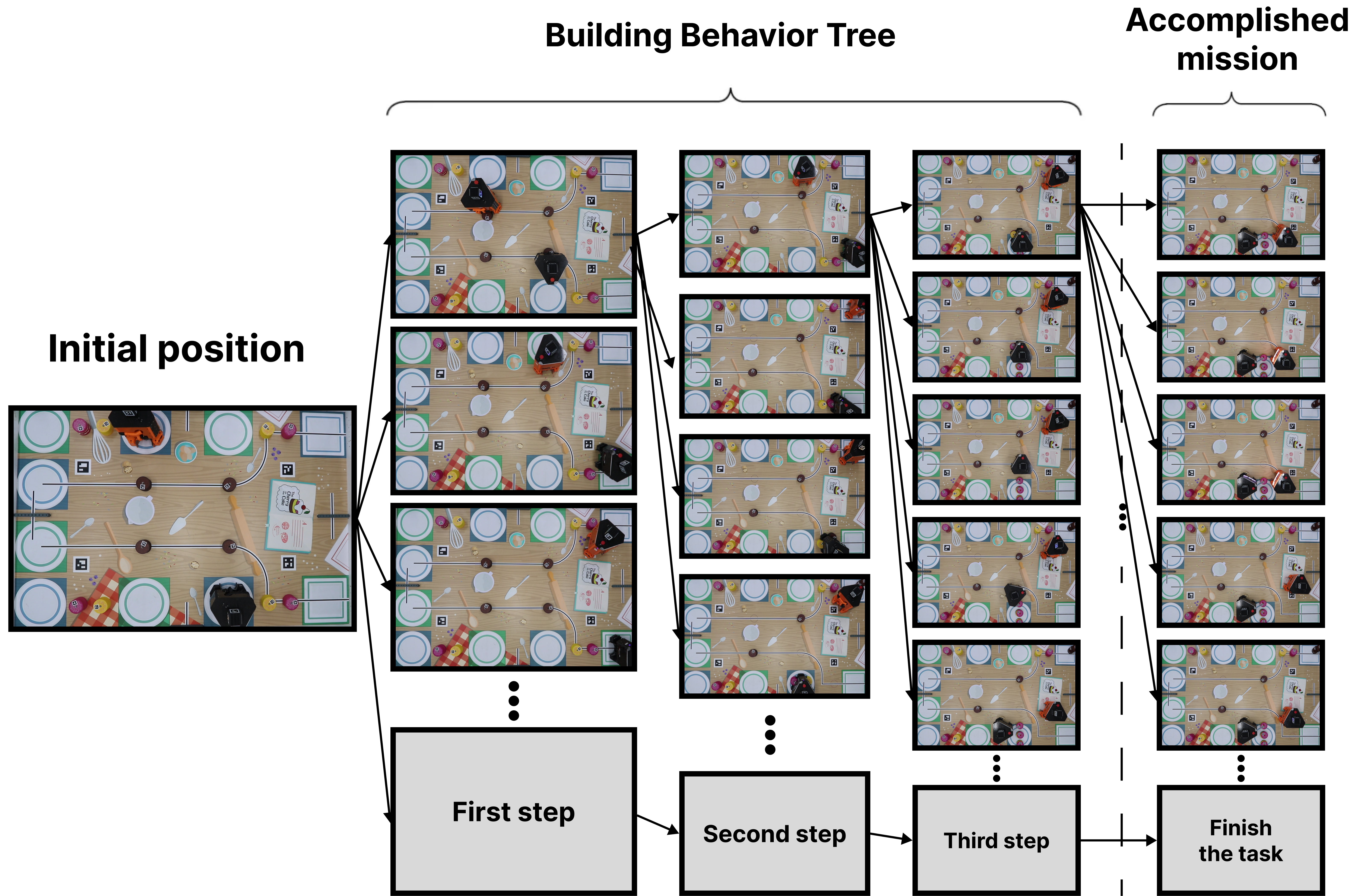}
    \caption[]{\small Strategy generation process. User defines the tasks, Large Language Model generates a Behavior Tree 
for robots to autonomously solve the tasks given the environment.}
    \label{fig:strategies}
\end{figure}

In this study, we employ the LLM-based BT generation approach that entails the utilization of a fine-tuned variant of the Falcon 7B LLM for BT generation. At the time of constructing our system, this model demonstrated exceptional metrics as per the LLM ranking \cite{lib:beeching2023open}. Employed as a proof of concept, this model served its purpose. Nevertheless, the relentless influx of newly released models presents an ongoing challenge to keep abreast of the latest developments in the field. 

The model is fine-tuned using a dataset generated with the text-davinci-003 model developed by OpenAI. To collect the dataset, pairs of commands for the robot and their corresponding BTs were generated using the ChatGPT API, while manually written BTs were used as examples for the model. Moreover, as our aim is to evaluate the performance of a multi-agent robot system when employing LLM under conditions that closely resemble real-world scenarios, experiments involving physical robots were conducted. 

A highly promising and readily apparent domain for the aforementioned application of dynamically generated BTs lies within the realm of robotics autonomous competitions. One such example is Eurobot, an esteemed international amateur robotics competition, hosted by Planète Sciences France. The primary technical challenge presented in Eurobot involves the construction of an autonomous robot or a pair of robots functioning in tandem. These robots are required to demonstrate reliability while effectively responding to actions initiated by their adversaries.

In the context of the 2023 iteration of the competition, the participating robots were tasked with executing a specific sequence of actions. These actions encompassed the collection and categorization of object sets, referred to as “cakes”, based on their respective colors. Additionally, the robots were required to gather and tally spherical objects, commonly referred to as “cherries”. Subsequently, the acquired objects needed to be transported and placed in designated locations on the playing field. Furthermore, the robots had to skillfully navigate the playing field to reach their assigned team positions, while simultaneously forecasting the accrued points resulting from the collected objects. These intricate conditions presented within the competition serve as an exceptional testing ground for the real-world evaluation of generating robotic behavior trees.

Thus, in our study we focus on the application of a multimodal LLM to control a tandem of mobile robots under conditions that closely emulate real-world scenarios. The process of generating strategies is depicted of Fig. {\ref{fig:strategies}}. The core modules of the model specifically designed for this purpose include a behavior generation module in a BT format and a module for discussing task outcomes.

\section{Related Works}
In recent times, the transformer models have gained significant popularity, becoming increasingly prevalent in various domains since the concept was introduced in the groundbreaking paper “Attention is all you need” by Vaswani et al. \cite{lib:vaswani2017attention}. These transformer models have shown remarkable performance in tasks such as language modeling, translation, and speech recognition, among others.

In particular, LLMs, which are based on the transformer architecture, have received tremendous attention in recent years. The rise in popularity has been facilitated by the emergence of models such as OpenAI's GPT3.5 and GPT4, whose features are discussed by Chen et al. \cite {lib:chen2023gpt35} and Bubeck et al. \cite{lib:bubeck2023gpt4} respectively. The introduction of ChatGPT by OpenAI \cite{lib:openai2022introducing} has significantly contributed to the growing popularity by bringing accessibility to GPT to a broader audience, taking the level of popularity to new heights. 

For a while, only those who collaborated with OpenAI had the opportunity to actively participate in the development of GPT models and witness their remarkable successes. However, the landscape changed with the advent of GPT-like LLM counterparts such as Google T5 described in paper Raffel et al. \cite{lib:raffel2020exploring} and Meta's LLaMa introduced in \cite{lib:touvron2023llama}, which offered alternative options that were not restricted to control of a single entity. Such models have gained widespread usage across various NLP applications, encompassing language generation, question-answering, sentiment analysis, and more. 

Furthermore, the emergence of open LLMs, including Stanford Alpaca \cite{lib:taori2023alpaca} model, has marked a turning point in LLM research. Work of E. J. Wang \cite{lib:wang2023flanalpacalora} offers a comprehensive resource for individuals seeking to fine-tune Alpaca 7B model using the identical methodology employed by Stanford University. It encompasses all the necessary components and instructions to facilitate the fine-tuning process. Unlike other LLMs, the Stanford Alpaca model not only became accessible to researchers worldwide LLM performing close to GPT3, but also possessed characteristics that enabled it to run even on average power personal computers. The model's architecture allows for fine-tuning to perform instruction-following tasks, which opens up enormous possibilities for researchers in related technical fields. As time went on, other, better performing open LLMs with the possibility of additional training, such as Falcon7B\cite{lib:penedo2023therefinedweb}, which we utilize in our work, began to appear. And lately the ranking of the best performing models \cite{lib:beeching2023open} is updated every month presenting better and better models both for 7B scales such as LLaMa 2 \cite{lib:touvron2023llama2} and Mistral \cite{lib:jiang2023mistral}, as well as more impressive ones like Falcon180B \cite {lib:2023falcon180}. Such models showcasing competitiveness with existing open-source chat models. The embodiment exhibit equivalent competency to certain proprietary models on evaluation sets examined, albeit trailing behind other advanced models like GPT-4.

Another particularly fascinating research direction involves the integration of LLMs, such as GPT, for applications in robotics \cite{lib:koubaa2023rosgpt}. These models have demonstrated the ability to generate coherent and contextually appropriate text, making them well-suited for a wide range of natural NLP tasks. In particular, they possess the ability to break down complex natural language tasks into elementary sub-tasks that can be executed by a robot \cite{lib:singh2023progprompt, lib:zhao2023chat}. Furthermore, LLMs have been employed to develop human-robot interaction systems leveraging NLP techniques \cite{lib:ding2023task, lib:huang2022inner}. Notably, the RT2 model \cite{lib:zitkovich2023rt2} has been introduced to tackle challenging manipulation tasks, while Tesla recently unveiled the Tesla Bot Optimus \cite{lib:su2023teslabot}, a humanoid robot equipped with advanced manipulation skills. The fusion of sophisticated LLMs with robotic systems represents a promising avenue that holds the potential for facilitating more effective collaboration between humans and robots both in highly specialized areas and for everyday tasks \cite{lib:ichter2022do}. 

However, only limited research has been conducted on the utilization of LLMs for generating robot behavior tree \cite{lib:cao2023robot, lib:wu2023embodied}. Previously, various approaches such as finite state machines \cite{lib:balogh2018usingfinite, lib:hussain2019finite}, Petri nets \cite{lib:lima1998petri, lib:ziparo2011petri, lib:lacerda2019petri, lib:lv2023optimal}, and BTs \cite{lib:colledanchise2017behavior, lib:colledanchise2018behavior, lib:ogren2022behavior, lib:colledanchise2021ontheimplementation, lib:safronov2020task} were commonly employed and still are being actively used to delineate the higher-level behavior of robots. Among these, BTs present an advanced means of specifying complex robot behavior \cite{lib:ghzouli2023behavior}. Traditionally, the application of such methodologies necessitates the involvement of highly qualified specialists with expertise in the specified domains, thereby rendering implementation challenging. However, the utilization of LLMs for generating behavior trees could be a game changer. 

Leveraging natural language processing techniques, LLMs for robot behavior, on the other hand, hold the potential to significantly enhance the adaptability and flexibility of AI-powered robotic systems. Y. Cao et al.'s work \cite{lib:cao2023robot} endeavored to create robot behavior through the utilization of LLM. The authors, however, employed an OpenAI product to populate BTs with behavior nodes. OpenAI products lack embodied experience in generating behavior trees, as they were not specifically trained for the purpose. Consequently, researchers imposed a limitation by fixing the behavior tree structure, restricting the use to sequence and action nodes only. While this approach imposes constraints on the modular structure advantages of BTs, the outcomes remain promising. Nevertheless, fine-tuning the LLM specifically for the task at hand usually yields much better results. This approach is anticipated to produce diverse BTs without any constraints on their structure.

Furthermore, several laboratories and industries with access to LLMs have been diligent in developing a methodology for building robot behavior that use, once again, GPT or its counterparts. One such work by Driess et al. \cite{lib:driess2023palme} focuses on the development of PaLM-E, an embodied multimodal language model that makes a robot to follow  human instructions in natural language while analyzing its environment. Another work by Brohan et al. \cite{lib:brohan2022rt1} presents a robotics transformer for real-world control at scale. Both of them are large models that require significant computational resources and training data. Beyond that, Boston Dynamics has also been exploring the potential of GPT for building robot behavior. They have successfully trained a robot dog Spot to provide operational reports to generate responses to inquiries, analyzing the surroundings \cite{lib:bostonspot}. However, it is employed only for vocal communication and navigation within the building without interactions with physical objects. 

Nevertheless, the generation of a diverse and extensive dataset with various BTs for robots of different structures and applications is necessary for fine-tuning LLM to achieve generation of complex BTs. While the approach of using Reinforcement Learning of from Human Feedback (RLHF) for LLM, as discussed by Stiennon et al. \cite{lib:stiennon2022learning}, is widely acknowledged, generating a substantial dataset would necessitate employing specialists who possess expertise in constructing BT. This, in turn, renders the task highly challenging. In our study, we enhance the methodology introduced by Y. Wang et al. \cite{lib:wang2022selfinstruct} as employed in the research conducted by Taori et al. \cite{lib:taori2023alpaca}. This approach involves the utilization of a text-davinci-003 model employing a self-instruct style to generate a dataset. The resultant model exhibits superior performance in certain tasks compared to the model responsible for dataset generation, owing to its fine-tuning tailored for these specific challenges.

The integration of LLMs and other transformer-based models in the field of robotics holds significant promise for improving human-robot interaction. The utilization of transformer-based LLMs enables remarkable advancements in addressing a wide range of tasks, including the comprehension of natural language commands and the decomposition of complex tasks into manageable subtasks.

\section{Advantages of Behavior Trees Approach}
\label{bt_for_robot}

\subsection{Advantages of Using Behavior Trees in Robotics}
BT is a hierarchical structure used to represent robot tasks at an abstract level, offering an alternative to the state machine paradigm. BT as an approach to constructing robot behavior is discussed by Colledanchise and Ögren \cite{lib:colledanchise2018behavior}. Formally, a BT is a directed rooted tree with leaf nodes responsible for task execution and branch nodes defining the control-flow logic. The leaf nodes are either Action or Condition nodes. The former specifies a primitive task and returns a Success signal when the task is completed. The latter is used to evaluate a Boolean condition, such as the satisfaction of a specific sensor reading. The most commonly used branch nodes are the Sequence and Fallback nodes. The Sequence node executes its child nodes sequentially until the first Failure signal is received. The Fallback node, on the other hand, executes its child nodes sequentially until the first Success signal is received. Using these four node types, a BT can achieve the same task execution as a state machine. But it has a lot of advantages, as modular structure allows to add, remove and replace nodes without having to reconstruct all structure.

BTs have become a widely used approach in robotics due to their intuitive and efficient control of robotic systems. They provide a structured way to represent and control the behavior of autonomous agents, making them well-suited for a wide range of robotic applications. They has been successfully applied in many robotics competitions and challenges, including the DARPA Robotics Challenge, RoboCup, and Eurobot. For example, in the DARPA Robotics Challenge, teams used BT to control their robots in tasks such as driving a car, opening doors, and using power tools, which was considered in the paper of Colledanchise and Ögren \cite{lib:colledanchise2014howbehavior}. In RoboCup, BT was used to control multi-robot systems in tasks such as soccer playing and search and rescue scenarios discussed by Safronov et al. \cite{lib:safronov2020task}. In Eurobot, it was used to control autonomous robots in tasks such as navigating a maze and manipulating objects considered in the paper of Granosik, et al. \cite{lib:granosik2016using}.

\subsection{Advantages of Using Behavior Trees as Output of LLM}
BTs, a hierarchical and modular structure, offer a promising solution for the development of transformer-based LLM. The structure enables the replacement of nodes with identical types. Ability to replace tokens make the structure well-suited for use as an output of the transformer. Additionally, the modularity and scalability of the BT structure allow for the easy addition of new nodes or modification of existing ones.

Another compelling feature of the BT structure is its option to use a subtree as part of a main tree. This means that generated BTs can be added to the node library and used as a part of more complex behavior. By operating in recursive mode, the model can build the BT first at the top level of abstraction and then descend to a lower level of abstraction, generating missing nodes from simpler components. This method allows for the creation of large and complex behavior structures while remaining within token length limitations in the model output. This approach positively impacts the model's performance and removes the limitations on the final size of the BT.

\section{Multimodal LLM to Multiagent Robot Control}

Current applications of LLM in robotics become strategically advantageous when addressing intricate sequences of actions with LLM-based decision making. These applications of LLMs lack significance when dealing with complicated systems where robots engaged in repetitive and monotonous tasks. As a long-term goal, we praise the creation of a universal AI-driven multi-agent robot system capable of understanding the goals of the task and solve it using existent agents.

Our approach differs significantly from traditional methods. We propose a solution in which a Language Model-Based Artificial Intelligence system takes control of an entire multi-agent system by generating behavior trees for these agents. Our system enables user interaction and control of robotic agents through dialogue. It comprises robotic agents and a multimodal LLM engaged in task discussions, command reception, and query responses. The LLM constructs BTs tailored for distinct agent types, thus facilitating the collaborative execution of intricate tasks. Agents autonomously execute these generated BTs and periodically update the system with information regarding their environment and current results. This BT generation approach, developed within our ISR Lab, provides a unique advantage in robot control. The system's ability to work seamlessly with different numbers and types of robots renders it highly scalable. In contrast to conventional methods that rely on a LLM solely to generate responses to external changes, our approach focuses on the generation of complex robot behaviors that consider a wide range of possible scenario evolutions. By creating a BT once, it inherently encompasses checks for multiple conditions and instructions on how to respond to them. This approach significantly reduces the number of requests to the LLM, effectively employing a single “brain” and enhancing resource efficiency.

    In practical scenarios, a multi-agent approach is often essential. In the field of robotics, individual machines perform straightforward tasks within their predefined behavior boundaries, while the collective system addresses intricate challenges. This phenomenon is readily observable in warehouse robotics, where robots carry out basic actions but collaborate to manage complex logistical demands. Similarly, in scientific drone exploration, each drone adheres to a simple plan, yet their collective endeavors result in detailed maps of remote terrains. Urban delivery robots excel in their individual roles of cargo transport, but the system's complexity lies in coordinating city-wide goods delivery. In facility inspection using multi-agent systems, robots follow predefined behaviors, while a centralized system ensures comprehensive coverage of the entire enterprise. Even in manufacturing, where individual robot tasks may involve repetitive manipulations, the entire conveyor system efficiently manufactures technological products. Hence, this is the typical scenario, with numerous robots contributing their simple components to a broader task. Consequently, providing each of these robots with its independent AI is an overly ambitious undertaking. 
    
    While existing examples of multi-agent systems are already in active use, their high-level control typically relies on a group of human operators or predefined programs tailored to specific tasks. Consequently, any task modification necessitates a team of experts and a significant amount of time to reconfigure the entire system. The crux of our proposed method is to delegate this high-level control to artificial intelligence. Our approach allows the rapid reorganization of the system to accommodate new tasks. To achieve this, it is only necessary to instruct the LLM-based AI manager about the tasks to be performed, which it will then execute using the available agents.

\subsection{Multimodal LLM to Control Eurobot Robots}

For prototyping the system within the laboratory, we turned to another task that has all the necessary qualities to realize the technology. Eurobot serves as an appropriate testbed for implementing LLM in a robot. Eurobot is an esteemed international robotics competition that provides a perfect platform for testing and evaluating robotic capabilities. The competition uses a multi-agent system of robots that can autonomously perform a given behavior. There are clear rules of the game and the criterion of victory, which is the goal of the system, but the changing conditions of the playing field generate thousands of scenarios. The competition challenges participants to design and operate autonomous robots that demonstrate reliability and responsiveness to adversary actions. In 2023 year, robots were tasked with collecting and sorting colored object sets known as 'cakes' and tallying spherical objects called 'cherries'. These objects then had to be transported and placed accurately within designated areas on the competition field. Thus, our robots, Black Samurai and Orange Shogan, were chosen as pioneers for LLM integration. Introducing LLM into a robot with initially designed High Level, Low Level, and Mechanics for different purposes poses a formidable challenge. The fact that the robots were initially created with support for strategies in the form of Behavior Trees assisted us in this endeavor. The objective was to launch a multimodal model in a manner that allows human interaction, while the generated BTs functioned seamlessly, just like those manually scripted before.
    
The hardware of our robot did not include components capable of supporting the autonomous execution of LLM with its 7 billion parameters. LLM was launched on a remote server accessible through SSH requests from the robot. To facilitate remote LLM execution, a server with Intel Broadwell featuring NVIDIA Tesla V100 hardware was leased. Natural language queries were sent to the server, and responses and strategies were sent back to the robot. Thus the whole system consisting of a multi-agent robotics system and a multimodal LLM was put together.

\section{System Overview}

In this section, we present an intricate analysis encompassing both the software and hardware aspects of the robotic system used for integration of the multi-modal LLM.

Through a meticulous exposition of the interplay between all robotic components and software parts, this section serves as a comprehensive guide towards not only understanding the intricacies of our achievements but also as a practical guide to facilitating autonomous replication of our results within the realm of creating a LLM-driven Multi-agent Robotic system.

\subsection{Mechanical Construction Overview}

\begin{figure}[!th]
    \centering
    \begin{subfigure}[b]{0.45\textwidth}
        \centering
        \includegraphics[width=\linewidth]{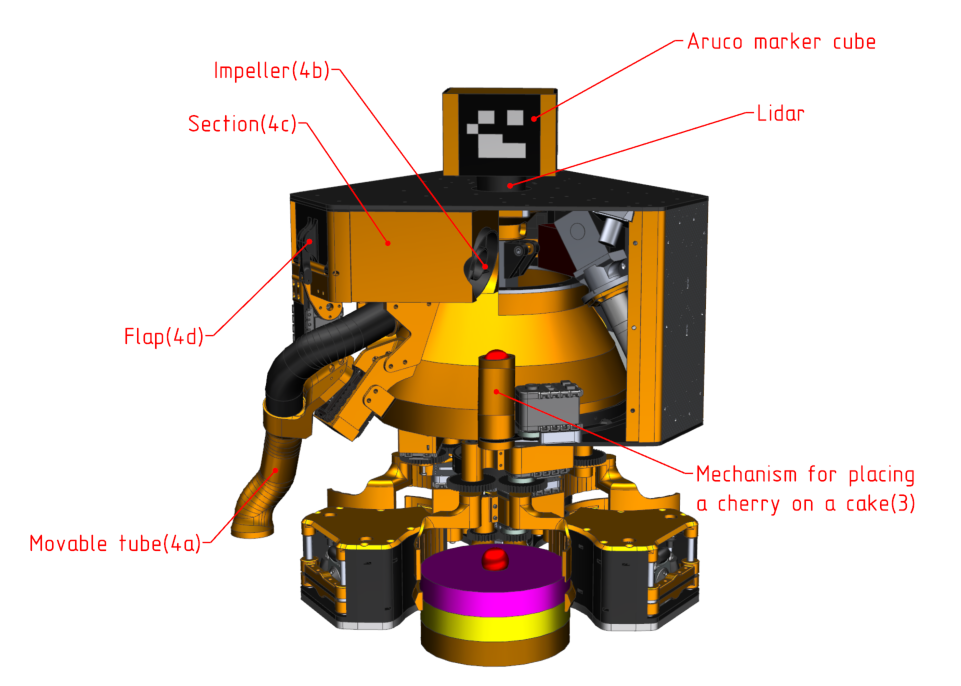}
        \caption{ }
        \label{fig:main_view}
    \end{subfigure}
    
    \vspace{1em}
    
    \begin{subfigure}[b]{0.45\textwidth}
        \centering
        \includegraphics[width=\linewidth]{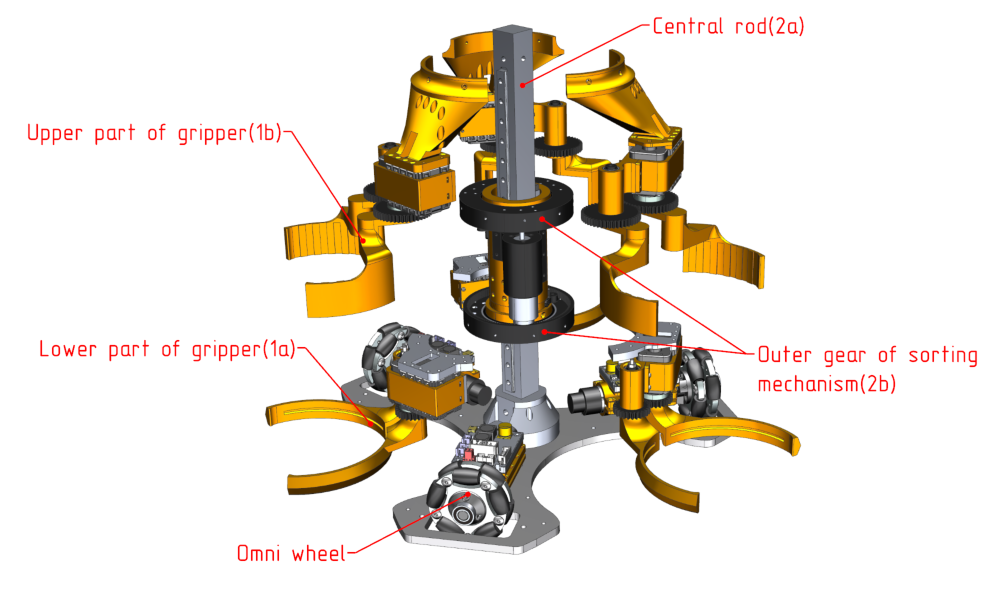}
        \caption{ }
        \label{fig:straddling_view}
    \end{subfigure}
    
    \caption[]{\small 3D CAD models of the robots. (a) Isometric view. (b) Straddling view of the frame and grippers.}
    \label{fig:constriction_views}
\end{figure}

The robots (Fig. \ref{fig:constriction_views}) are specifically crafted to adhere to the regulations of the Eurobot competition. The steel frame attaches the essential elements of each robot, which moves on a platform featuring omni-wheels and a suspension system. Within the mechanical structure, key components include grippers, a sorting mechanism, a cherry dispenser, and a cherry collection mechanism.

Each of the three grippers are divided into a lower part (1a) for gripping the bottom cake layer and an upper part (1b) for gripping the two upper cake layers. All sections of the gripper are controlled by a servomotor and attached to the outer gears of a sorting mechanism. The sorting mechanism consists of lifting and rotating parts and is raised and lowered on a central rod (2a) by a motor through a belt. The rotating part comprises two parts, resembling a planetary gear system, with one input gear rotating the outer gear (2b) using a motor. The cherry dispenser is composed of a special tube and a dedicated flap controlled by a servomotor. The actuation of the flap facilitates the controlled release of cherries onto the cake. The cherry collection mechanism comprises distinct elements: a specially shaped movable tube (4a) regulated by two servomotors; two impellers (4b) designed for the suction of cherries from a designated section (4c) within the robot and subsequently expel them into a specialized receptacle; and outer and inner flaps (4d) under the control of servomotors.

\subsection{Embedded Systems Overview}

Robotic electronics serve as a crucial intermediary, facilitating seamless interaction between the high-level computer and the set of onboard devices, including the omni-wheel platform, actuators, switches, and sensors. Its paramount function extends to providing the high-level side with abstraction layers and interfaces, thereby streamlining the overall interaction process.

\begin{figure}
    \centering
    \includegraphics[width=\linewidth]{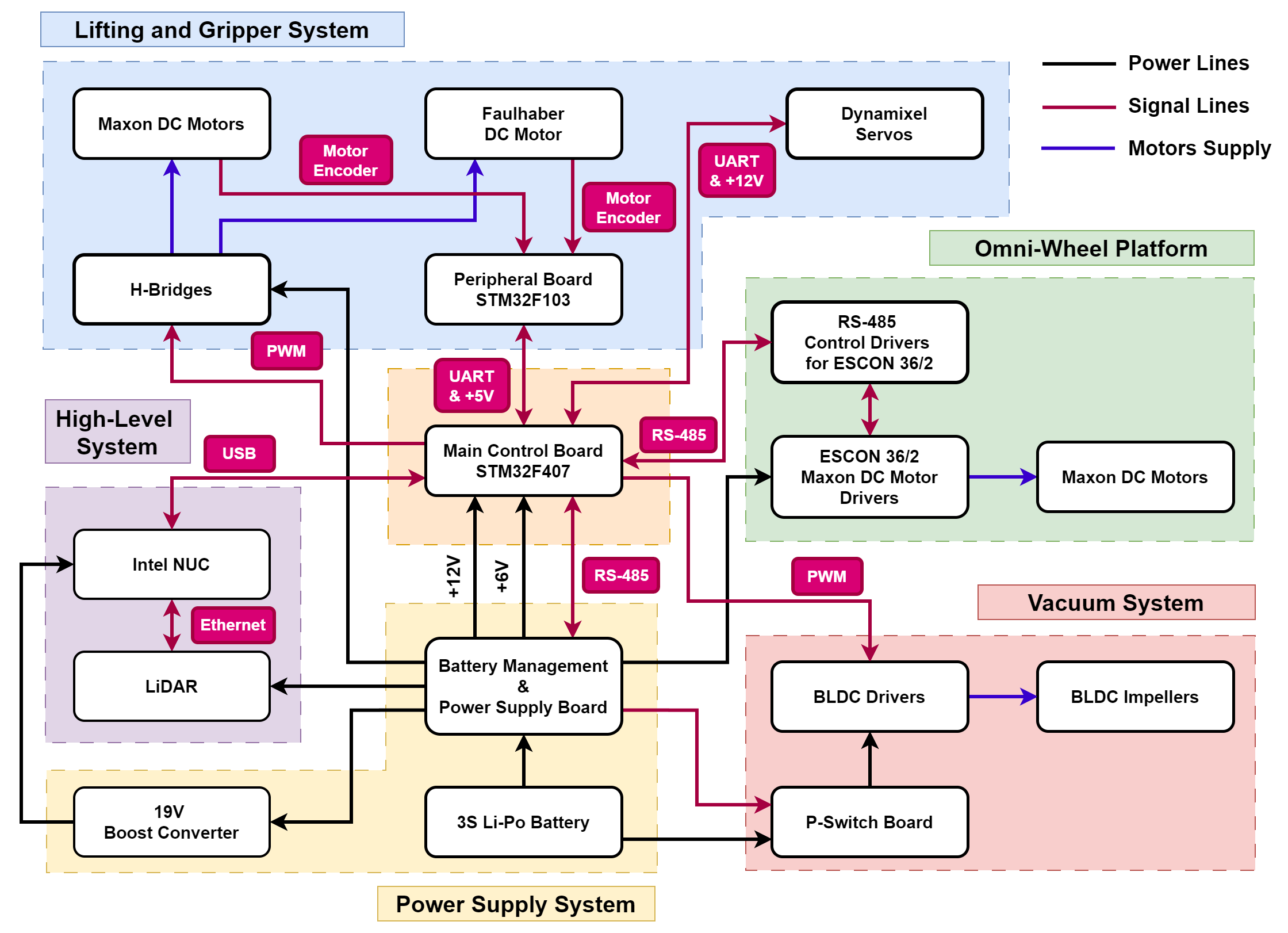}
    \caption[]{\small Block diagram of the robot electronics.}
    \label{fig:electronics_scheme}
\end{figure}

Essential time-critical operations and low-level control algorithms are housed within the electronic components, further solidifying its role in the system. Additionally, the electronics adeptly oversees the management of the onboard battery, ensuring its safe and optimal functioning, while effectively distributing power to all pertinent robot components. An illustrative representation of the electronics' structure is presented on Fig. {\ref{fig:electronics_scheme}}.

At the heart of the robot electronics lies the main control board, centered around the STM32F407VG Microcontroller Unit (MCU). This pivotal board draws power from both the +5V and +12V power lines, while the remaining electronic entities are deemed downstream and subservient, operating under the firm control and connection of this central board. Furthermore, the high-level computer serves as the upstream side, establishing connectivity to the main control board via a USB cable.

\subsection{High-Level Systems Overview}

The high-level system is designed to manage data acquired directly from sensors such as LIDAR and cameras, as well as from the embedded system, particularly wheel encoders. Furthermore, it facilitates interaction with robot actuators through various commands, empowering developers to devise sophisticated strategies. To fulfill these requirements, the system is built upon the ROS2 Humble framework \cite{lib:macenski2022robot}, which runs on an Intel NUC computer installed with Ubuntu 22.04. The computer communicates with STM32 via USART interface. Therefore, this module receives commands and sends responses through ROS topics and performs serialization or deserialization to match the USART data exchange format. Besides that, the module publishes wheel odometry to ROS topics with frequency of 40 Hz. 
    
The localization module is important in accurately determining a robot's position within a playing field. We employed a particle filter (PF) methodology \cite{lib:gustaffson2002particle}, integrating data from wheel encoders, LiDAR, and the computer vision (CV) system. Initially, the module computes odometry measurements from individual sensor data, referencing the robot's previous pose. For the extraction of LiDAR-based odometry, we incorporated a laser scan matcher module, which operates on the principles of the Canonical Scan Matcher \cite{lib:censi2008anicpvariant}. These odometry measurements are pivotal during the “predict” phase of the particle filter. It is worth noting that each sensor possesses a weighted value, determining the proportion of particles influenced by their respective measurements. The raw LiDAR point cloud serves as the primary source for landmark detection. These landmarks are situated at the vertices of a triangle beyond the edges of the playing field and are characterized by cylindrical shapes encased in retroreflective material. The detection procedure combines a sequence of operations: filtering points based on light intensity return, outlier removal, and landmark center identification via optimization techniques. Subsequently, data association is executed, correlating the detected centers with the global positions of the landmarks. Concluding the process, the pinpointed landmark centers become crucial during the “update” phase of the PF. This facilitates the synthesis of measurements, resulting in a refined and filtered pose determination for the robot.
    
The CV system is designed to perform two main functions: game object detection and robot localization. The system consists of a Jetson Xavier AGX microcomputer and an Imagine Source DFK 33UX250 Global Shutter camera equipped with a Computar fisheye lens. The camera is mounted on a tower-type tracking device and has an overhead view of the playground. The first aspect of the system involves accurate detection and identification of game objects, specifically cakes. This is achieved using the Hough algorithm \cite{lib:leavers1989adynamic} for circle finding, effectively tracking all cakes on the field. Additionally, the system can determine the color and number of layers of each detected cake. The second functionality excels in precisely localizing robots on the field. Robots are equipped with cubes containing AruCo markers on their faces, which are designed to be visible from any angle, ensuring reliable detection even under varying perspectives. Using a single AruCo marker, the system accurately localizes the robot in three-dimensional space. This feature is particularly valuable in dynamic scenarios involving multiple robots, where LiDAR localization may not provide accurate data. The CV system utilizes the OpenCV python library \cite{lib:bradski2000theopencv} for the detection and recognition of AruCo markers.
    
Navigation module is responsible for creating an environment map, generating safe trajectories for robots and controlling their velocities. Initially, the module consistently acquires refreshed coordinates of robots, encompassing those of opponents, as well as game objects, from the localization and computer vision (CV) modules. This data is used to generate an environment map. It is structured as a grid consisting of 1x1 cm cells, with each cell assigned values of 0 (indicating not occupied) or 1 (indicating occupied). Full map is made by adding occupied circles with a certain radius on a static map, which consists of field borders and fixed cherry holders. Subsequently, the module receives a target point from the BT module and formulates a secure and concise path from the current position to the target by employing the RRT* algorithm \cite{lib:karaman2011anytime}.
During each iteration of progressing towards the goal, the module examines whether opponent robots obstruct the established path. In the event of such obstruction, the module generates a new path from the current position. If the target proves unreachable, a stop command is issued along with feedback to the BT module.Ultimately, once the path is established, the module generates a smooth velocity profile using the minimum jerk algorithm \cite{lib:bianco2013minimumjerk} and sends velocities commands to STM32. Velocities are limited according to a minimum distance to other robots to avoid collisions with them. To hold on a path two PID regulators are used: one for trajectory and one for angle. 
    
BT module provides interface to create human-readable strategies for robots. For this purpose behavior-tree-cpp-v3 library \cite{lib:faconti2019behaviortreecpp} is used. Each command for actuators is wrapped into a certain action with input and output ports. Additionally, helpful conditions are provided, such as “If-Then-Else” or bounds for execution time for an action. Strategies are written as xml-files.

\section{LLM Training Methods and Recipes}

The primary objective of our research was to train a model capable of constructing BTs for robots in multi-agent system based on natural language commands from operators. To achieve this, we fine-tuned the Falcon model containing 7 billion parameters using the Low-Rank Adaptation PEFT method \cite{lib:hu2021lora} on a dataset generated with the text-davinci-003 model. The PEFT methods allows efficiently and productively fine-tune the LLM without having to retrain the entire model, as discussed by G. Pu et al. \cite{lib:pu2023empirical}. PEFT methods are particularly useful when tuning LLMs is too computationally expensive. Instead of tuning all the model's parameters, PEFT methods adjust only a small number of additional parameters, thus significantly reducing computational and storage costs. The PEFT method we employed to fine-tune the LLM, is the Low-Rank Adaptation (LoRA) method, which adds low-rank matrices to transformer layers and adjusts only those matrices instead of the entire model. The application of the LoRA method to the LLM is described by E. J. Hu et al. \cite{lib:hu2021lora}. The aforementioned Falcon 7B was selected for our study because it demonstrated superior performance in terms of open model rating \cite{lib:beeching2023open} at the time of system development.

\subsection{Behavior Tree Generation Module Fine-tuning Process}

This subsection delves into the process of structuring a dataset tailored for training a LLM for the generation of BT. The LLM sequentially generates words, and its training dataset comprises robot's behavior as prompts and corresponding complete answers. Thus, for our purpose each dataset sample consists of an instruction for building the robot's behavior and an output represented as a logically and structurally correct BT that allows the robot to execute the task. The instruction includes a system prompt that is common to all samples. This part is necessary for further work when the LLM will not only build BTs but also perform other tasks. After the unchangeable part of the instruction, there is a description of the required robot behavior. When generating this part of the samples using the text-davinci-003 model, special attention was paid to the naturalness of the request. It should be a simple and logically understandable command that a human could give, as it is the human who will give the command to the robot when using the fine-tuned model. The generated BT consist of Action and Condition nodes that the robot can perform. In addition to Action and Condition nodes, the BT sometimes contains SubTree nodes. These nodes work like Action nodes, but themselves are compiled BTs. Adding SubTree nodes allows generating the overall logic of the robot's behavior in several stages by generating missing elements from the available nodes. This approach avoids generating large constructions using LLM in one go, which would significantly increase the required memory, as the need to store attention between all sequence elements leads to a quadratic increase in the required memory with the increase in sequence length. 
    
Furthermore, we proceeded with the dataset generation process. Initially, we possessed a list of nodes available for addition to the BT of our robots, and we aimed for the model to generate behaviors solely from these predefined nodes. To achieve this, we generated diverse samples of Behavior Trees using a Python script without resorting to LLM methods. The script produced a varied set of action combinations and parameters, along with corresponding descriptions following a specific template. While we retained the BTs in the dataset unchanged, we paraphrased the descriptions using the ChatGPT API. As a result, our dataset not only encompassed comprehensive and strategically valid behaviors but also exhibited request diversity. The training dataset comprised 7500 BT samples, each paired with its corresponding user command. The fine-tuning process lasted for three epochs on an Nvidia Tesla V100, totaling 10 hours.

\subsection{Question Answering Module Fine-tuning Process}

\begin{figure}
    \centering
    \includegraphics[width=\linewidth]{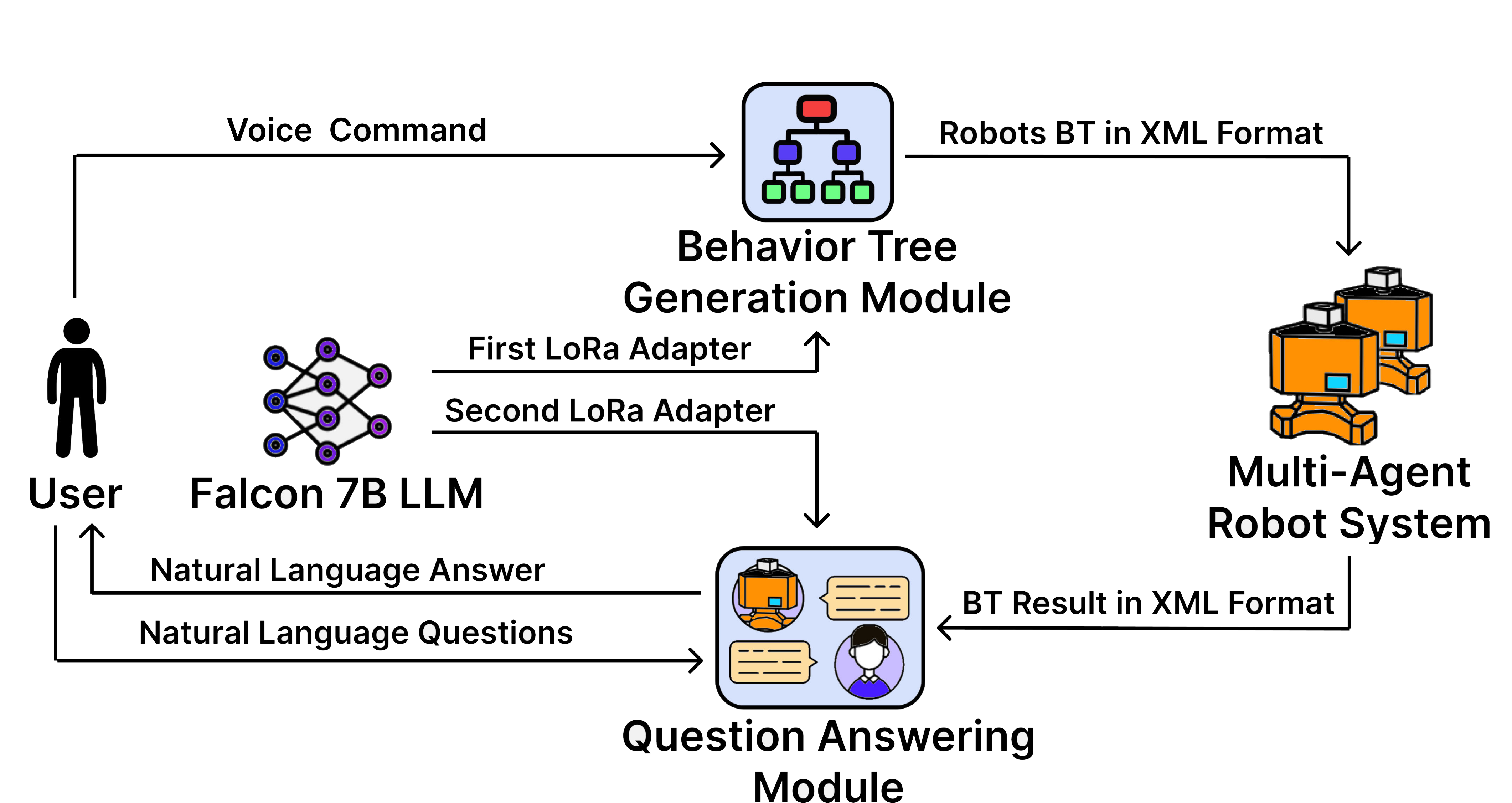}
    \caption[]{\small System architecture of LLM-MARS.}
    \label{fig:system_architecture}
\end{figure}

For effective human-robot interaction, it is insufficient for the robot to merely receive and execute commands. It must also provide feedback and answer to questions regarding its behavior. Considering our overarching goal of achieving autonomous AI-powered robots, it would be impractical to develop a separate full-fledged model to handle the task of answering questions. Operating two 7B models simultaneously on a single robot's microcomputer significantly elevates its hardware requirements. Consequently, we conceived the idea of using a single LLM alongside multiple LoRa adapters as a solution. This approach offers advantages in terms of compactness and resource efficiency. However, it also has drawbacks, such as the need to switch between adapters, taking approximately 40 seconds. We also explored the possibility of using one adapter for both tasks; however, this negatively impacted the correctness of the structure of the generated BTs, jeopardizing the robot's functionality. 
    
To address the generation of answers to questions about the outcome of the robot's behavior, we drew inspiration from the work of Boston Dynamics \cite {lib:bostonspot}. An XML file containing the context of the behavior's results is assembled based on the BT and data from control sensors. This file is then passed as input to the LLM along with the user's question to the robot. In original project, authors employed the ChatGPT API to generate responses for question-answering. In our research, we utilize the ChatGPT API to generate 11000 question-answering samples based on XML files in our specified format, each with various outcomes. After three epochs of training on Nvidia Tesla V100 on this dataset, taking 12 hours, we integrated a second LoRa adapter into our model, rendering it multimodal. The resulting architecture is presented on Fig. {\ref{fig:system_architecture}}.

At the moment we have developed two adapters. During further development, the number of adapters, and therefore the capabilities of the multimodal system, can be expanded.

\section{Experimental Evaluation}

\subsection{Experiment on Human Ability to Recognize Model-Generated Behavior Tree}
\label{experiment}

To assess the performance of the model, we conducted an evaluation of individuals' capacity to differentiate between LLM-generated BTs and manually authored ones. A similar investigation was conducted by Brown et al. \cite{lib:brown2020language} but with the distinction that in our study, all participants possessed knowledge regarding the underlying principles and construction of BTs.

\paragraph{Participants}

We recruited a total of 15 participants, comprising both undergraduate and graduate students specializing in the Robotics track, to assess the quality of BT generation. Among the participants, ten individuals were members of local Eurobot team and regularly worked with BT. The remaining five participants had no prior exposure to BT and received explicit instructions on the underlying principles and construction techniques of BT. Prior to the commencement of the experiment, all participants provided their informed consent, thereby confirming their familiarity with the operational and structural aspects of robot BTs.

\paragraph{Procedure}

To conduct the experiment, a set of 10 BTs was generated using LLM. These BTs were designed to simulate various actions and interactions that a mobile robot could perform. Additionally, a separate set of 10 BTs was created manually, aiming to achieve similar functionalities and behaviors as the LLM generated BTs. Participants were presented with descriptions of the robot's behavior and pairs of BTs, consisting of one LLM-generated BT and one human-created BT. The order of presentation was randomized to avoid any bias. Participants were then asked to evaluate the BTs and determine which one was created by the LLM and which one was human-created. They were instructed to rely on their subjective perception and any distinguishing features they could identify.

\paragraph{Experimental Results}

\begin{figure}
    \centering
    \includegraphics[width=\linewidth]{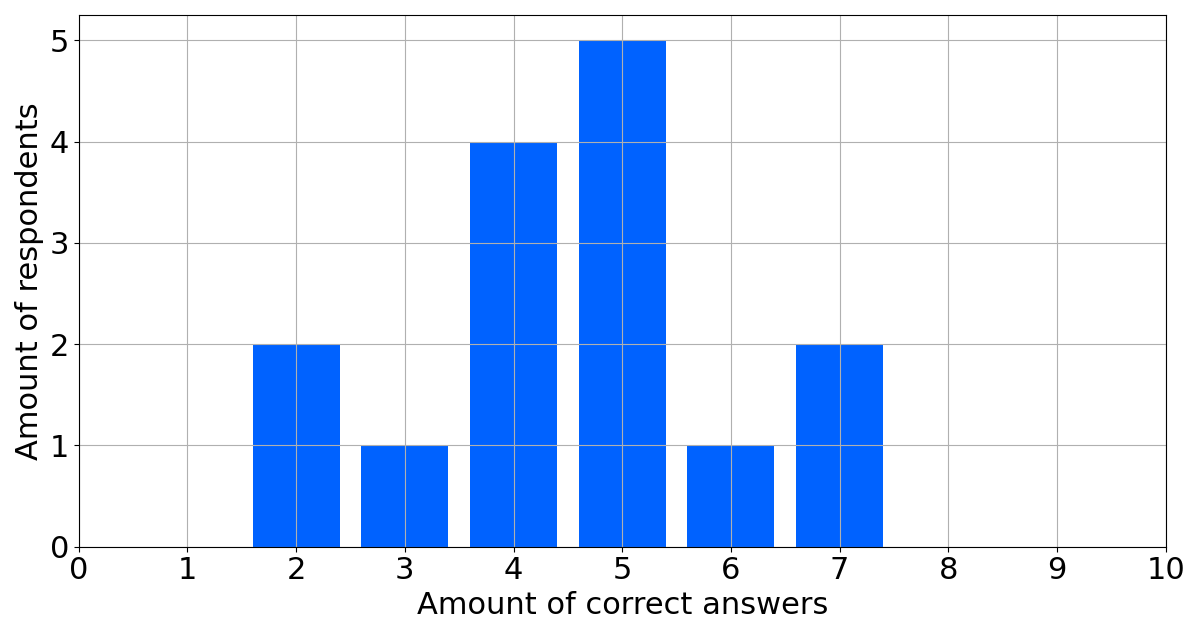}
    \caption[]{\small Correct answer distribution for experiment on human ability to recognize model-generated behavior tree.}
    \label{fig:experiment0}
\end{figure}

After the completion of the assessment phase for all pairs of BTs, the resulting data were collected and subjected to analysis. The findings of the experiment are visually depicted in Fig. {\ref{fig:experiment0}}. The mean score of 4.53 correct answers out of a total of 10 suggests that participants' ability to differentiate between BTs generated by the LLM and those created by humans was comparable to random chance. 

In order to examine potential correlations between survey responses and specific questions, a one-way analysis of variance (ANOVA) was conducted with a significance level set at 5\%. The analysis revealed no statistically significant difference in users' perceptions of different questions (F = 0.75, p = 0.66 $>$ 0.05). This suggests that there is no evidence to support the claim that the probability of a correct answer depends on the question. To determine whether there exist discernible distinctions between human-generated BTs and those generated by the LLM, a t-test was conducted with a significance level set at 5\% to evaluate the null hypothesis that they cannot be distinguished from each other. Under this null hypothesis, the average number of correct answers provided by the subjects would be 5 out of 10, and the average score would be 0.5. Based on our results (mean score = 0.453, t-statistic = -1.2, critical value = 2.145, p $>$ 0.05), we have no reason to reject the null hypothesis, which suggests that the mean score is not significantly different from 0.5.

The user study, involving 15 volunteers, yielded no substantive indications of subjective disparities between LLM-generated BTs and human-generated BTs. The LLM model demonstrates the capability to generate robot behavior with result close to human-created BTs, at least in terms of subjective perception within the context of our experiment.

\subsection{Behavior Generation Performance Evaluation}

After the experiment demonstrated the ability of LLM to generate BT for given tasks, we conducted experiment on a real multi-agent robot system. The experiment aimed to evaluate the extent to which the formed BTs reflect the tasks given. 

\paragraph{Participants}

The experiment involved the participation of authors of this work who assessed the LLM's ability to generate BTs for designated tasks.

\paragraph{Procedure}

\begin{figure}
    \centering
    \includegraphics[width=\linewidth]{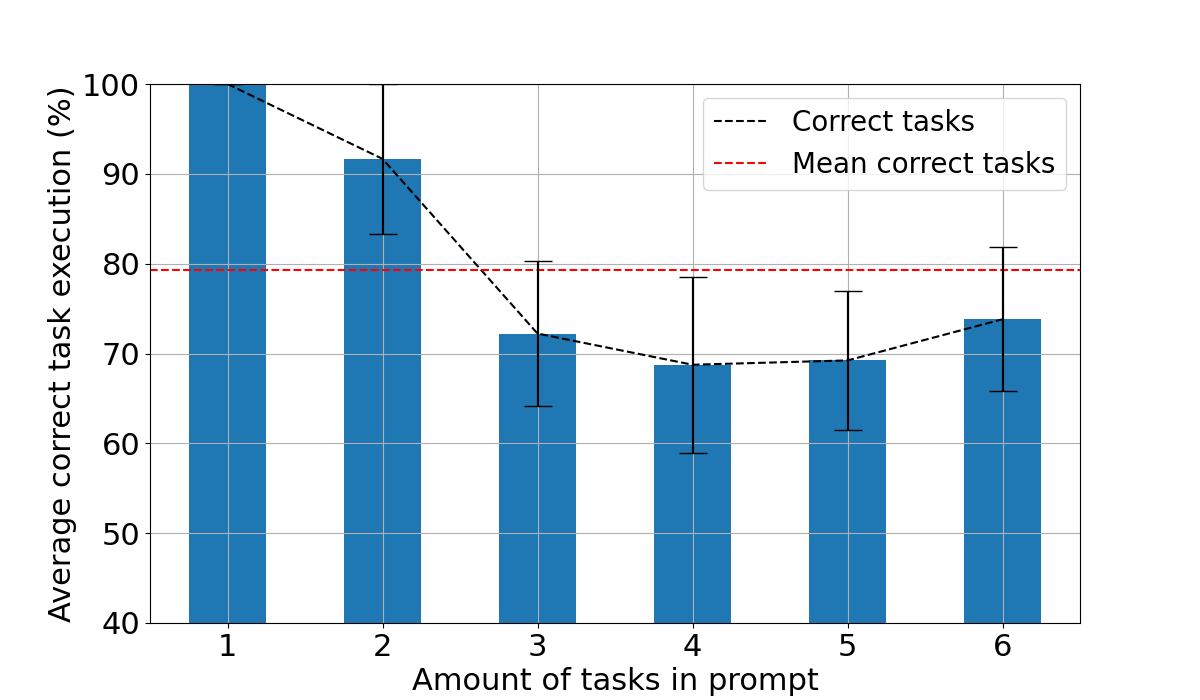}
    \caption[]{\small Task quantity's impact on Large Language Model's performance accuracy.}
    \label{fig:experiment1}
\end{figure}

The LLM was presented with sixty commands, each containing one to six tasks. For each task variant, ten examples were available, ensuring a comprehensive evaluation of task complexities. The primary focus was on analyzing the accuracy of integrating robot tasks into the BT. The result of the experiment is presented on Fig. {\ref{fig:experiment1}}.

\paragraph{Experimental Results}

The results of the experiment showcased the remarkable performance of the LLM in generating BTs based on commands. The LLM demonstrated an accuracy of 70.44\% in integrating provided robot tasks into the BTs. For commands with up to two tasks, integration accuracy exceeded 90\%. The average percentage of correctly added tasks within a command was found to be 79.28\%. Notably, task integration accuracy in BTs was influenced by the number of tasks in a command.

\subsection{Expert Evaluation of Question Responses}

\paragraph{Participants}

The following experiment involved the collection of 50 responses from the Language Model (LLM) concerning robot behavior. These responses were subject to evaluation by ten individual with specialized knowledge and experience in the field related to robot behavior based on three criteria: accuracy, relevance, and informativeness.

\begin{figure}
    \centering
    \includegraphics[width=\linewidth]{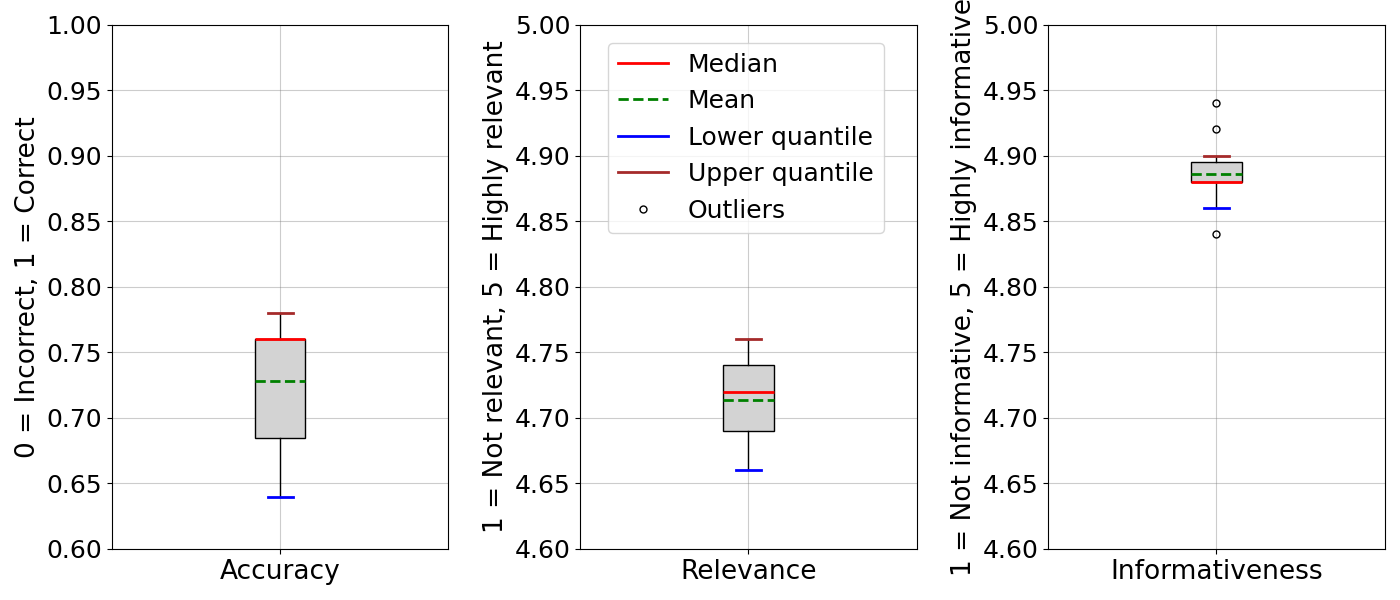}
    \caption[]{\small Average assessment of Large Language Model parameters.}
    \label{fig:experiment2}
\end{figure}

\paragraph{Procedure}

The responses from the LLM were evaluated by expert raters based on three criteria. Accuracy was scored as 0 for incorrect and 1 for correct responses. Relevance was rated on a scale of 1 to 5 (1 = not relevant, 5 = highly relevant), and informativeness was assessed on a scale of 1 to 5, with higher scores indicating more relevant information. To validate the significance of the data, a Krippendorff's Alpha Reliability test was conducted to assess expert agreement. The corresponding data is presented on Fig. {\ref{fig:experiment2}}.

\paragraph{Experimental Results}

The average accuracy of the LLM was 72.8\%, with relevance and informativeness scores equaling 4.71 and 4.89, respectively.

To validate the significance of the data, a Krippendorff's Alpha Reliability test was conducted to assess expert agreement. The results of Krippendorff's Alpha Reliability test are the following:

    \begin{itemize}
        \item Reliability for Accuracy: The Krippendorff's alpha equals 0.8239, demonstrating a robust agreement among evaluators regarding accuracy.
        
        \item Reliability for Relevance: The Krippendorff's alpha equals 0.7229, indicating a moderate level of agreement among evaluators regarding relevance.
        
        \item Reliability for Informativeness: The Krippendorff's alpha equals 0.7594, also signifying a moderate level of agreement among evaluators regarding informativeness.
    \end{itemize}

\section{Results and Discussion}

The results (Fig. {\ref{fig:experiment1}}) demonstrated outstanding performance of LLM in generating BT based on commands. LLM accurately integrated 70.44\% of all provided robot tasks into the generated Behavior Trees. For commands with up to two tasks, the integration accuracy exceeded 90\%. Additionally, the average percentage of correctly added tasks within a command equals 79.28\%.
Moreover, task integration accuracy in BT was influenced by the number of tasks in a command. A one-way analysis of variance (ANOVA) with a significance level of 0.05 confirmed the statistical significance of this relationship (F-statistic = 2.61, p-value = 0.035).
Specifically, LLM performed better with commands containing fewer tasks. The performance slightly improved when dealing with commands with up to 6 tasks, likely due to the fact that this amount of tasks make strategy similar to strategies for Eurobot that was used to generate the dataset for LLM finetuning.

The results of question response evaluation for each expert (Fig. {\ref{fig:experiment2}}) indicated a strong performance of the LLM in addressing questions related to robot behavior. The average accuracy was 72.8\%. The relevance and informativeness score equal 4.71 and 4.89 respectively.  

The results of Krippendorff's Alpha Reliability test indicated a strong agreement among the expert evaluators for the Accuracy parameter. However, for the Relevance and Informativeness parameters, the level of agreement was moderate. This disparity can be attributed to the more subjective nature of relevance and informativeness compared to the Accuracy.

In our research, we explore the exciting fusion of LLM and robotics. We developed first technology to create a multi-agent robotic system capable of engaging in dialogue with humans, constructing complex behavioral strategies for its agents based on human commands, and providing feedback on task execution. This technology has been put into practice in a multi-agent robotic system under near real-world conditions. Experimental results show that robots correctly execute given compound commands at an average of 79.28\%, with higher accuracy (exceeding 90\%) observed for commands containing one or two tasks. Furthermore, expert evaluation of question responses revealed that the answers provided by the system exhibit high accuracy, relevance, and informativeness.

Although the results obtained during the research fully justified our expectations, this technology still offers a vast expanse for further exploration and refinement. In future endeavors, our focus will be on enhancing the command processing capabilities of the model, particularly for commands comprising a larger number of tasks. This improvement can be achieved through a two-step approach involving command decomposition into simpler tasks and generating BTs based on these tasks. The decomposition stage could be realized by adding a dedicated function to the BT generation adapter or training a separate adapter specifically for this purpose.

\section{Conclusion}

In conclusion, our development and implementation of LLM-MARS mark a groundbreaking advancement in the convergence of artificial intelligence and robotics. This innovative approach, utilizing a Large Language Model based on the Falcon 7B, introduces a paradigm shift in robot control, enabling the management of multi-agent robotic systems through dynamic dialogues.

Our trials within the Eurobot 2023 game rules demonstrate significant success, with an impressive average task execution accuracy of 79.28\%. Particularly noteworthy is the system's proficiency in integrating tasks into Behavior Trees, showcasing the effectiveness of the employed Large Language Model used in such a scenario. The observed correlation between task integration accuracy and the number of tasks in a command suggests targeted enhancements for command processing.

Expert evaluations further affirm the reliability and potential of LLM-MARS in practical scenarios, emphasizing high accuracy, relevance, and informativeness in responses to questions. These findings also illuminate areas for future research, particularly in enhancing the system's handling of more complex commands through advanced techniques like command decomposition and specialized adapters.

The core achievement lies in our pioneering approach to robot management, allowing a swarm of robots to possess AI capabilities under the control of a single LLM. This strategy is especially crucial for scenarios where robot groups collaboratively tackle complex tasks, each responding to simple instructions. E.g. mobile robot groups could revolutionize logistics by fully replacing human warehouse personnel. Additionally, groups of robots or drones could autonomously carry out exploration missions, searching for valuable artifacts and conducting terrain research. Furthermore, the application of this technology to create a multi-agent system of manipulator robots represents a step forward in the direction of the collaboration between humans and machines, also knows as Industry 5.0.

The primary advantage of our approach is that the LLM configures the system for specific tasks, eliminating the need for specialized human programming for each robot. This not only streamlines the process but also empowers the AI as a system manager, capable of answering inquiries about its performance.

\begin{IEEEbiography}
[{\includegraphics[width=1in,height=1.25in,clip,keepaspectratio]{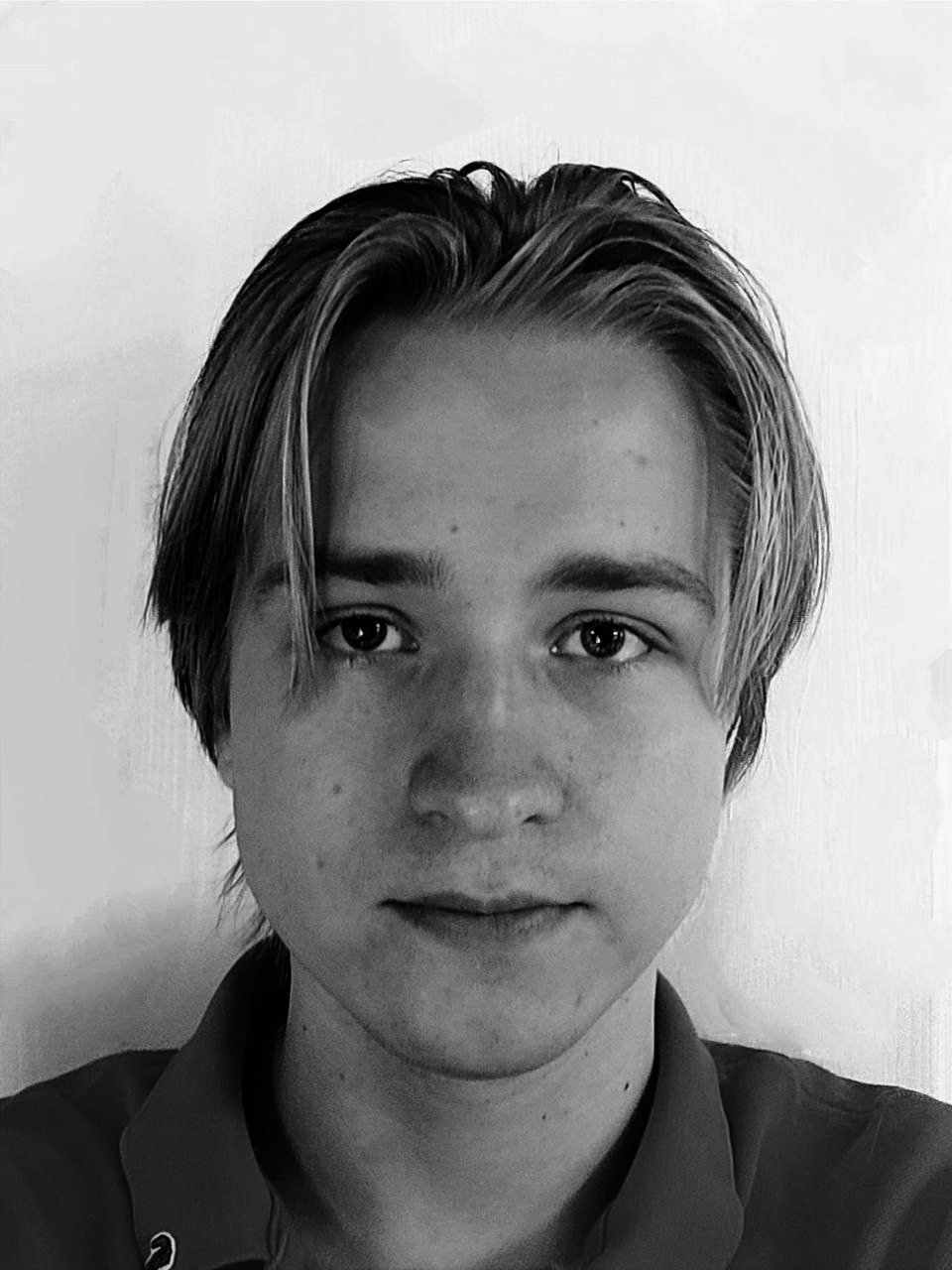}}]{Artem Lykov}
graduated with honors from the Bauman Moscow State Technical University in 2022 and received the B.Sc. degree in Robotic Systems and Mechatronics. He is currently pursuing his Master's degree at the Skolkovo Institute of Science and Technology, in the Engineering Systems department, Robotics track. He conducts research and development in the field of LLM-based robots with AI in the Intelligent Space Robotics Laboratory. His research interests include AI, LLM, multi-agent robotic systems, autonomous robots, human-robot interaction and haptics. Artem received Best Paper Award at Asia Haptics conference in 2022. As a member of the Reset team of the Skolkovo Institute of Science and Technology, in 2023 he won the champion title at the national stage of the Eurobot robotics competition.
\end{IEEEbiography}

\begin{IEEEbiography}
[{\includegraphics[width=1in,height=1.25in,clip,keepaspectratio]{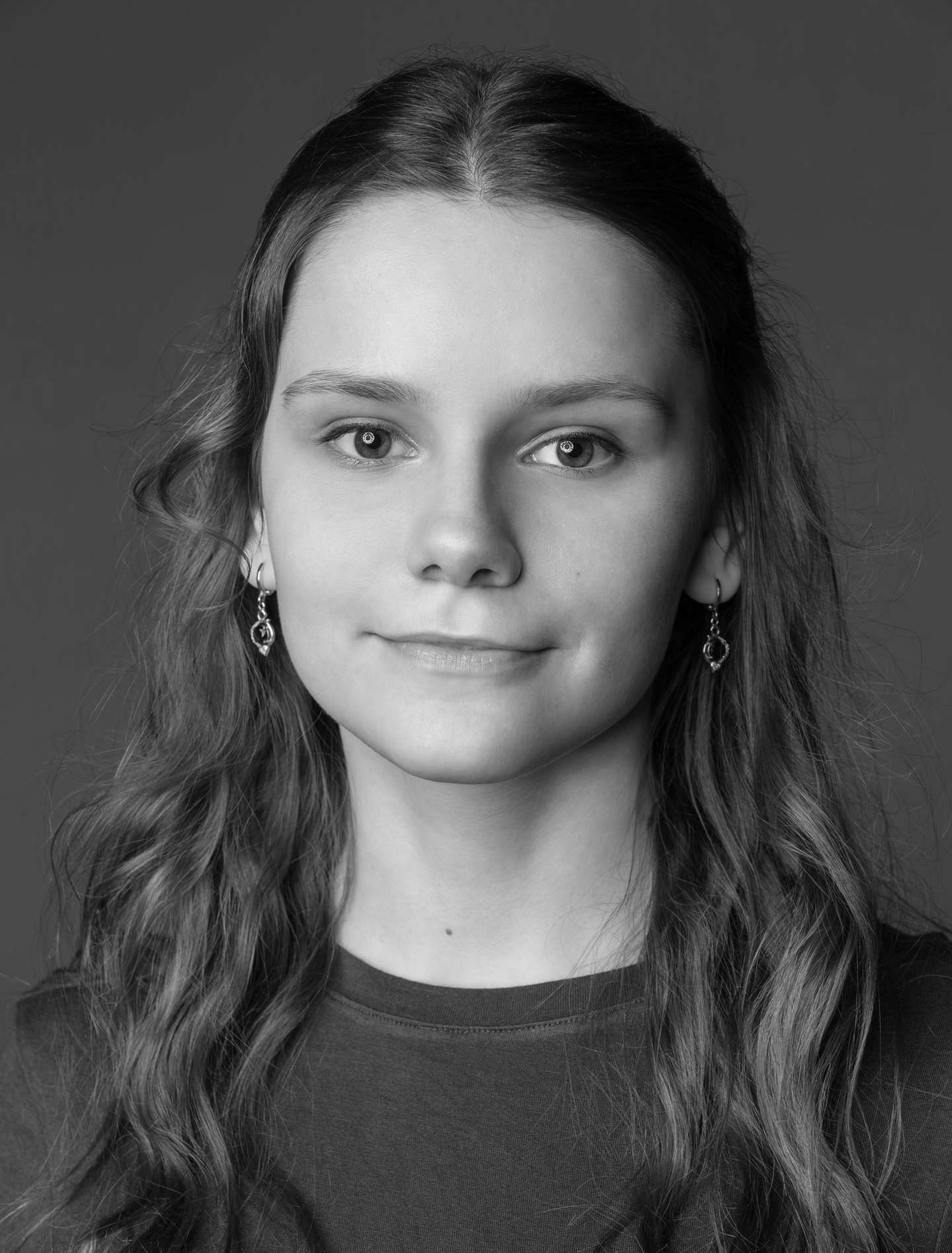}}]{Maria Dronova}  
graduated with honors from the Bauman Moscow State Technical University in 2022 and received the B.Sc. degree in Rocket and Space Technology. Enrolled in 2022 in the Master's program at the Skolkovo Institute of Science and Technology in the Engineering Systems department, Robotics track. Maria, as the member of ReSET Skoltech team, became the champion of Eurobot Russia 2023, prestigious autonomous robot competition. She is the author of 5 papers, her research interests lay in the field of drones, path planning algorithms and deep learning.
\end{IEEEbiography}

\begin{IEEEbiography}
[{\includegraphics[width=1in,height=1.25in,clip,keepaspectratio]{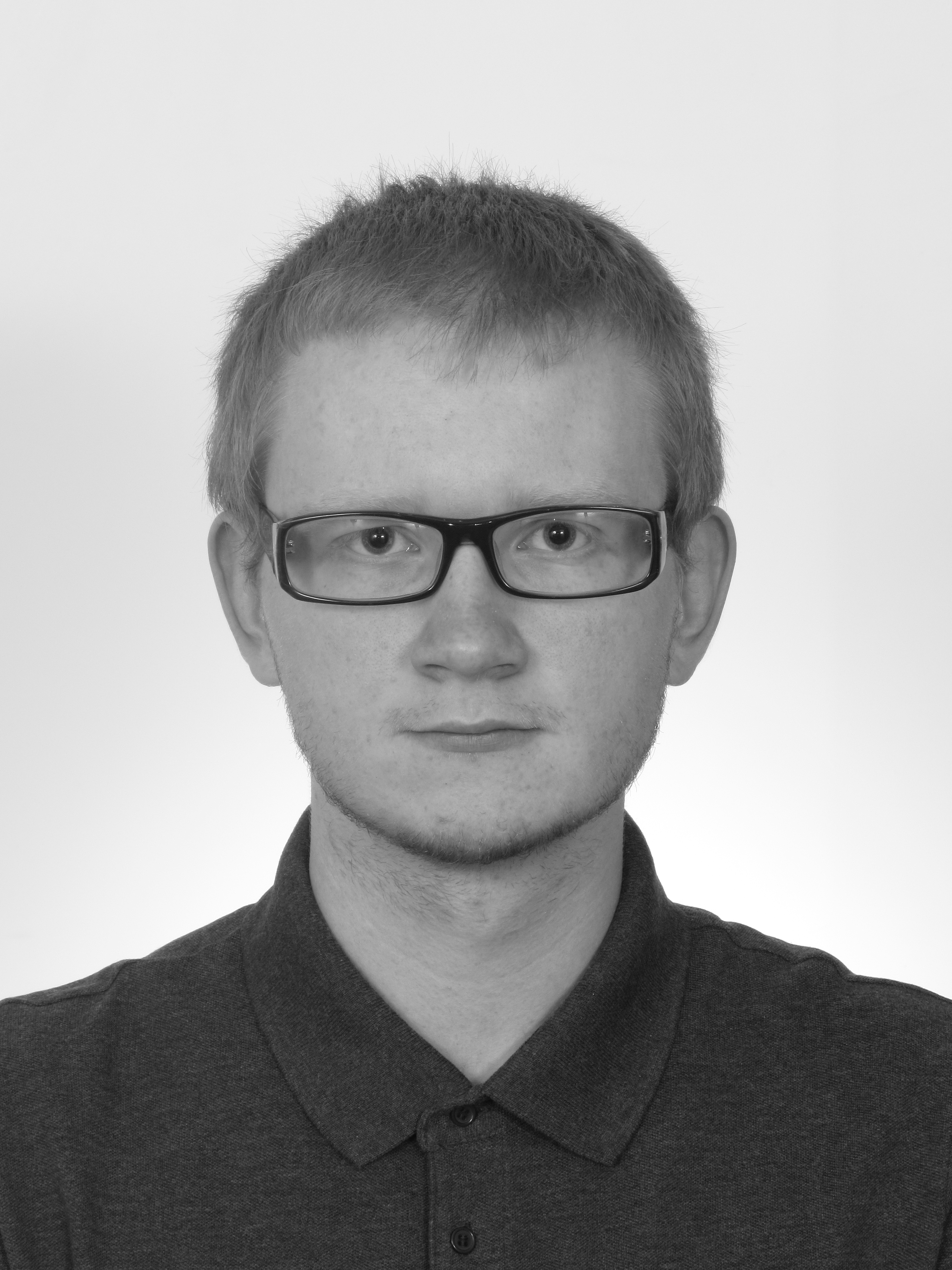}}]{Nikolay Naglov}
graduated with honors from the Saratov State Technical University in 2022 and received the B.Sc. degree in Mechatronics and Robotics. He is currently pursuing his Master's degree at the Skolkovo Institute of Science and Technology, in the Engineering Systems department, Robotics track. His research interests include navigation of autonomous mobile robots, human-robot interaction, AI and LLM. Nikolay is the champion of Eurobot Russia 2023 autonomous mobile robot competition as a member of RESET Skoltech team.
\end{IEEEbiography}

\begin{IEEEbiography}
[{\includegraphics[width=1in,height=1.25in,clip,keepaspectratio]{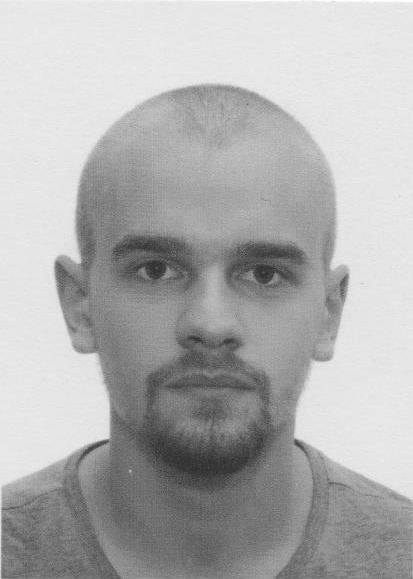}}]{Mikhail Litvinov}
received his Bachelor's degree in Radio Engineering and Computer Technology from the Moscow Institute of Physics and Technology. He is currently pursuing his Master's degree at the Skolkovo Institute of Science and Technology, in the Intelligent Space Robotics Laboratory. Mikhail was a part of the ReSET team that became the champion of Eurobot Russia 2023, respectable autonomous robot competition. His research interests include autonomous mobile robotics, human-robot interaction, computer vision, and deep learning.
\end{IEEEbiography}

\begin{IEEEbiography}
[{\includegraphics[width=1in,height=1.25in,clip,keepaspectratio]{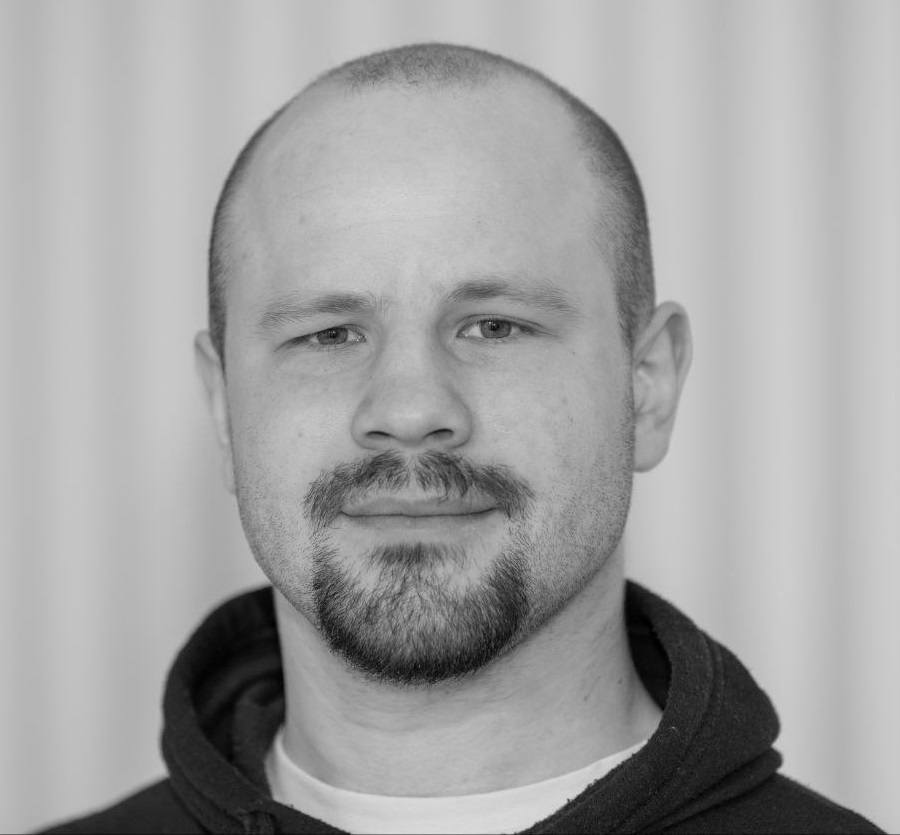}}]{Sergei Satsevich}
graduated from Bauman Moscow State Technical University in 2019 with an M.Sc. degree in Wheeled Vehicles. Subsequently, he worked in the aerospace industry as an mechanical engineer. Presently, he is pursuing a Master's degree at the Skolkovo Institute of Science and Technology in the Engineering Systems department, specializing in the Robotics track. His research interests encompass autonomous robots, legged robots, drones, soft robotics, control engineering, and reinforcement learning. Being part of the Reset team at the Skolkovo Institute of Science and Technology, he secured the championship title during the national round of the Eurobot robotics competition in 2023.
\end{IEEEbiography}

\begin{IEEEbiography}
[{\includegraphics[width=1in,height=1.25in,clip,keepaspectratio]{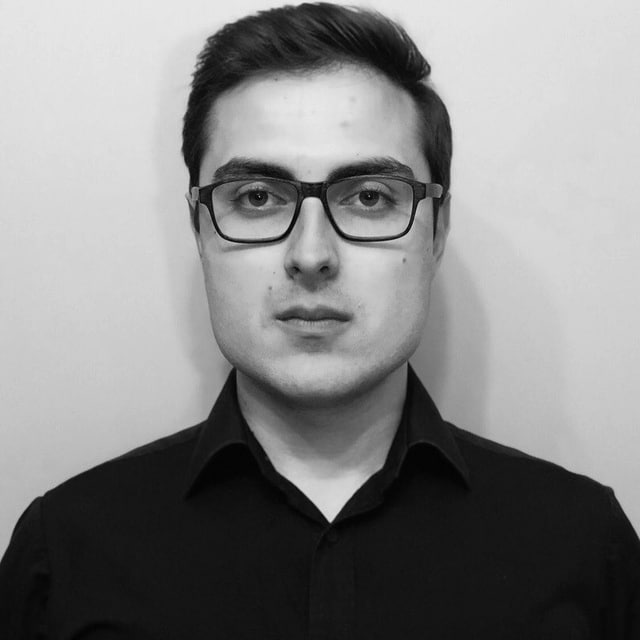}}]{Artem Bazhenov}
graduated from Gubkin University as a mechanical engineer, then gained work experience in the oil and gas industry and is now a master's student at Skoltech. He has created several mechanical design for devices, more than 15 different electronic devices of high complexity with PCB boards, and has a strong background not only in engineering, but also in Deep Learning.
His research interests include LLM-based robotics control, Reinforcement-Learning, human-robot interaction.
\end{IEEEbiography}

\begin{IEEEbiography}
[{\includegraphics[width=1in,height=1.25in,clip,keepaspectratio]{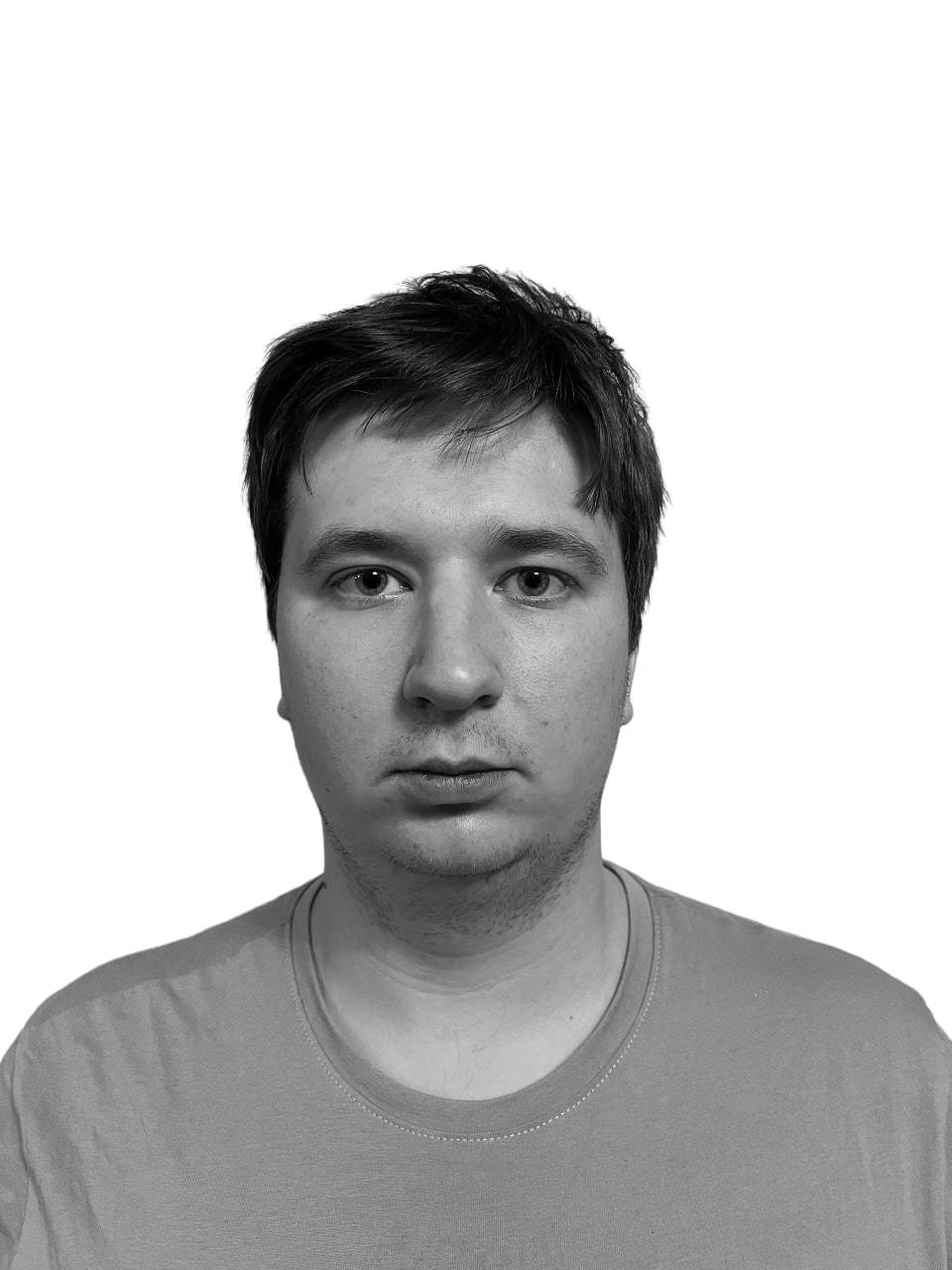}}]{Vladimir Berman}
graduated with honors from Bauman Moscow State Technical University in the field of Biomedical Technologies. Currently, he is pursuing a Master's degree in Robotics at the Skolkovo Institute of Science and Technology. His research activities encompass Deep Learning, Reinforcement Learning, Natural Language Processing and Robotics Transformers in multi-agent systems. Additionally, he has a background in electronics and backend development.
\end{IEEEbiography}

\begin{IEEEbiography}
[{\includegraphics[width=1in,height=1.25in,clip,keepaspectratio]{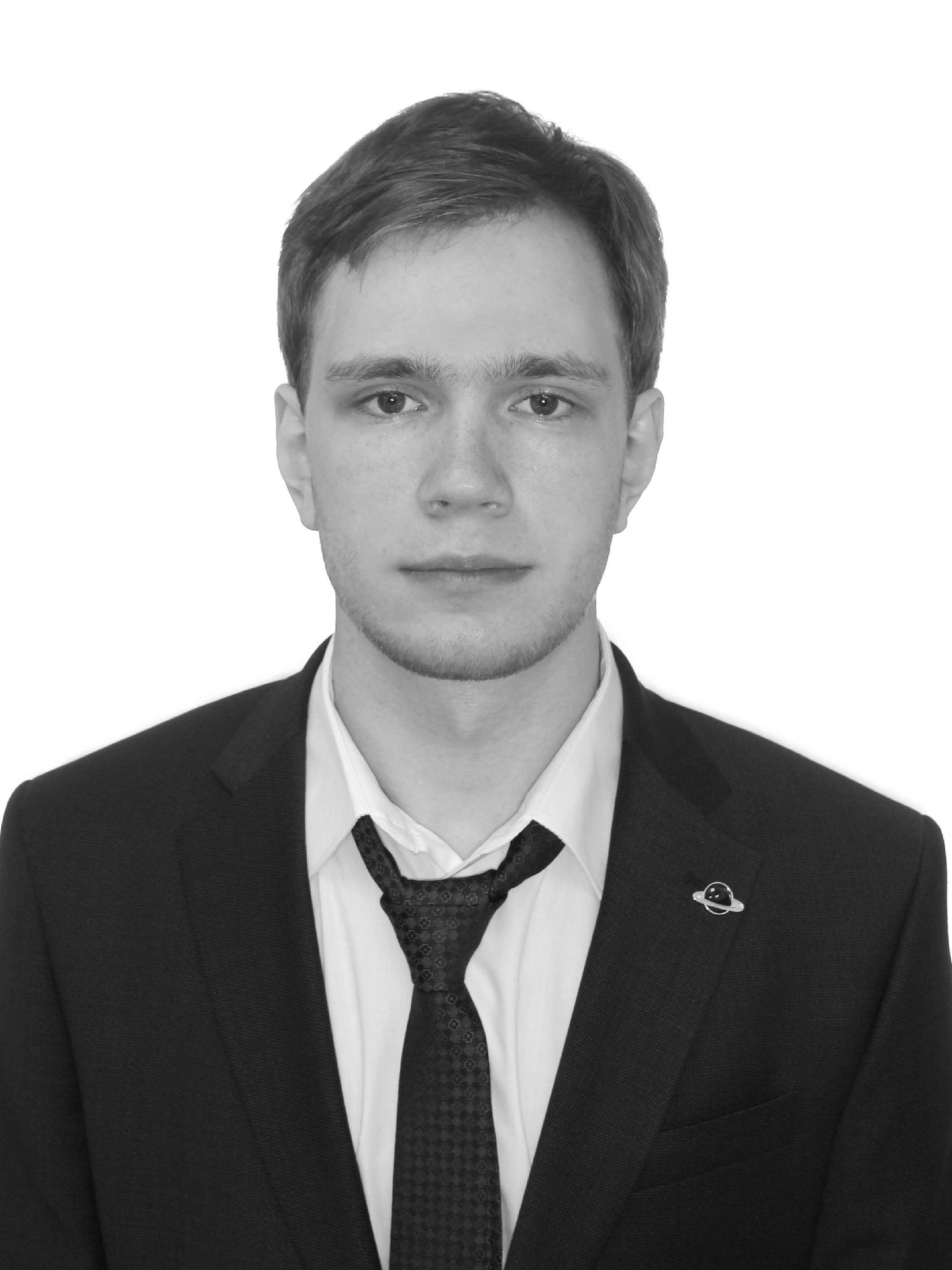}}]{Aleksei Shcherbak}
is a PhD student of Skolkovo Institute of Science and Technology (Skoltech), Russia. Aleksei received his MSc degree in Physics at Moscow State University, Physics department, Russia, in 2019. His research interests include wearable sensing, robotics, electronics development, firmware development, computationals and machine learning in medical related applications. Aleksei is the champion of Eurobot Russia 2022 and 2023 autonomous mobile robot competition as a member of RESET Skoltech team.
\end{IEEEbiography}

\begin{IEEEbiography}
[{\includegraphics[width=1in,height=1.25in,clip,keepaspectratio]{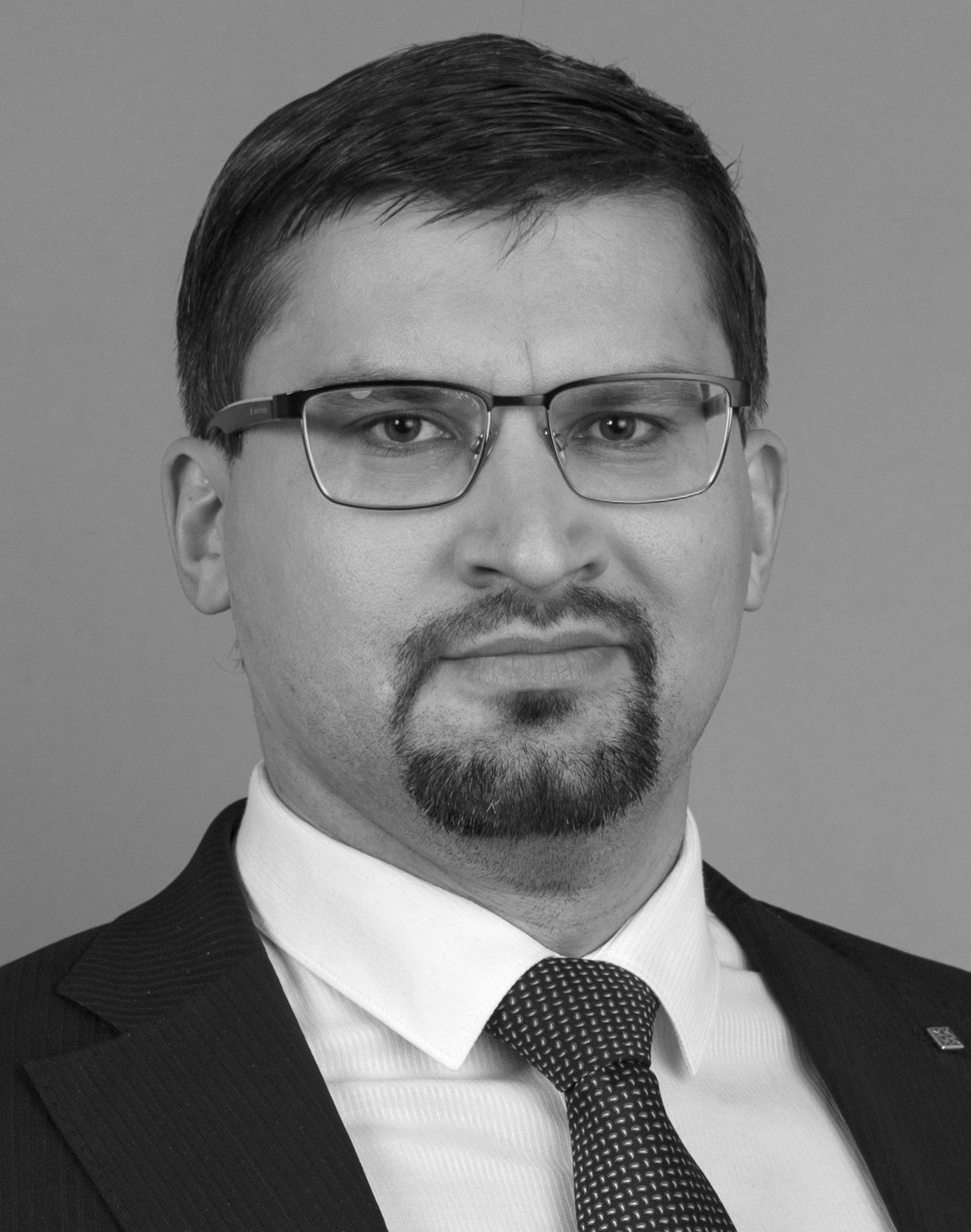}}]{Dzmitry Tsetserukou}
 received the Ph.D. degree in Information Science and Technology from the University of Tokyo, Japan, in 2007. From 2007 to 2009, he was a JSPS Post-Doctoral Fellow at the University of Tokyo. He worked as Assistant Professor at the Electronics-Inspired Interdisciplinary Research Institute, Toyohashi University of Technology from 2010 to 2014. From August 2014 he works at Skolkovo Institute of Science and Technology as Associate Professor and Head of Intelligent Space Robotics Laboratory. Dzmitry is a member of the Institute of Electrical and Electronics Engineers (IEEE) since 2006 and the author of 128 scientific papers indexed in Scopus, 10 patents, and a book.

 His research interests include autonomous robots, swarm of drones, LLM, AI,  human-robot interaction, haptic and tactile displays. In 2023 he was awarded by Elsevier (Scopus) the rank of top 2\% of the world’s most cited researchers. Dzmitry received such prestigious Awards as Best Paper Award (Asia Haptics 2022), Finalist of Franklin V. Taylor Memorial Award (IEEE SMC 2021), Best Demonstration Award (ACM SIGGRAPH Asia 2019), Best Demonstration Award (Bronze Prize AsiaHaptics 2018),  Laval Virtual Awards (ACM SIGGRAPH 2016), and a Best Paper Award (ACM Augmented Human 2010). Team ReSet of Skoltech supervised by Dzmitry became the 7-time Champions of Russia in autonomous robot competition Eurobot (2016-2023) and vice-champions of the Eurobot World in 2019, France.  
\end{IEEEbiography}

\end{document}